# ReportAGE: Automatically extracting the exact age of Twitter users based on self-reports in tweets

Ari Z. Klein[1*], Arjun Magge[1], Graciela Gonzalez-Hernandez[1]

[1] Department of Biostatistics, Epidemiology, and Informatics, Perelman School of Medicine, University of Pennsylvania, Philadelphia, Pennsylvania, United States of America

* Corresponding author

Email: ariklein@pennmedicine.upenn.edu

# Abstract

Advancing the utility of social media data for research applications requires methods for automatically detecting demographic information about social media study populations, including users' age. The objective of this study was to develop and evaluate a method that automatically identifies the exact age of users based on self-reports in their tweets. Our end-to-end automatic natural language processing (NLP) pipeline, ReportAGE, includes query patterns to retrieve tweets that potentially mention an age, a classifier to distinguish retrieved tweets that self-report the user's exact age ("age" tweets) and those that do not ("no age" tweets), and rule-based extraction to identify the age. To develop and evaluate ReportAGE, we manually annotated 11,000 tweets that matched the query patterns. Based on 1000 tweets that were annotated by all five annotators, inter-annotator agreement (Fleiss' kappa) was 0.80 for distinguishing "age" and "no age" tweets, and 0.95 for identifying the exact age among the "age" tweets on which the annotators agreed. A deep neural network classifier, based on a RoBERTa-Large pretrained transformer model, achieved the highest $F_1$-score of 0.914 (precision = 0.905, recall = 0.942) for the "age" class. When the age extraction was evaluated using the classifier's predictions, it achieved an $F_1$-score of 0.855 (precision = 0.805, recall = 0.914) for the "age" class. When it was evaluated directly on the held-out test set, it achieved an $F_1$-score of 0.931 (precision = 0.873, recall = 0.998) for the "age" class. We deployed ReportAGE on a collection of more than 1.2 billion tweets posted by 245,927 users, and predicted ages for 132,637 (54%) of them. Scaling the detection of exact age to this large number of users can advance the utility of social media data for research applications that do not align with the predefined age groupings of extant binary or multi-class classification approaches.

# Introduction

Considering that 72% of adults in the United States use social media [1], it has been widely utilized as source of data in a variety of research applications. However, a common limitation of this research is that users of particular platforms are not representative of the general population [2]. Thus, advancing the utility of social media data requires methods for automatically detecting demographic information about social media study populations, including users' age. Most studies have approached the automatic detection of age as binary classification [3, 4] or multi-class classification [5-9] of predefined age groups. These studies first identify the age of users based on their or other users' posts, their profile metadata, or external information, and then evaluate the prediction of the users' age group based on modeling a large collection of their posts [3-6], their profile metadata [7], a combination of their posts and profile metadata [8], or their followers and followers' friends [9]. While the automatic classification of age groups may be a suitable approach for specific demographic inquiries about social media users, the fact that the number and range of the age groups vary across the studies suggests that this approach is not generalizable to all applications.

Automatically identifying the *exact* age of social media users, rather than their age groups, would enable the large-scale use of social media data for applications that do not align with the predefined groupings of extant binary or multi-class models, such as identifying specific age-related risk factors for observational studies [10], or selecting age-based study populations [11]. Nguyen et al. [5] have developed a regression model for automatically identifying the exact age of Dutch Twitter users, but their evaluation was based in part on annotations of perceived age, which may be influenced by humans' systematic biases and, thus, different from the users' actual age [12]. For this reason, the annotators in Nguyen et al.'s [5] study were asked to assess how confident they were in identifying exact age, with a margin of error of up to ten years. Sloan et al. [13] have developed rules to automatically identify self-reports of Twitter users' actual age, but their high-precision approach extracts age only from users' profile metadata, and was able to automatically detect age for only 1,470 (0.37%) of 398,452 users.

The objective of this study was to develop and evaluate a method that automatically identifies the exact age of users based on self-reports in their tweets. Individual tweets have been used to manually verify self-reports of age for evaluating prediction models, but, to the best of our knowledge, methods to extract self-reports in tweets have not been scaled for automatically identifying users' age. For example, Al Zamal et al. [3] have identified users' age by searching for self-reported birthday announcements in tweets (e.g., *happy ##(st|nd|rd|th) birthday to me*), but then used the ages to evaluate a binary model that infers an age group from 1,000 of the users' most recent tweets, based on linguistic differences associated with age. Similarly, Morgan-Lopez et al. [8] have identified users' age by searching for self-reported birthday announcements in tweets, but then used the ages to evaluate a multi-class model that infers an age group from 200 of the users' most recent tweets and their profile metadata. These high-recall approaches can potentially infer an age group for any user, since they do not rely on explicit reports of age. For this same reason, however, they are not designed for identifying users' exact age, which limits their application beyond predefined groupings.

In this paper, we present ReportAGE (Recall-Enhanced Pipeline for Obtaining Reports of Tweeters' Ages Given Exactly), situated in the gap between rule-based (high precision) and predictive modeling (high recall) approaches. ReportAGE utilizes individual tweets as a resource to identify exact age for a large number of users, overcoming the sparse reports of age in users' profiles. A tweet-based approach, however, does present challenges in natural language processing (NLP). Query patterns that have high precision within the constraints of a user's profile (e.g., *years old*) would return significantly more noise from tweets, and, while a small number of patterns may capture most of the ways in which users express their age in a profile, tweets afford a wider range of expressions. These expressions may require deriving the user's age from references to the past or future, whereas ages in profiles are likely to refer to the present. To address these challenges, we have designed ReportAGE as an end-to-end NLP pipeline that includes high-recall query patterns, a deep neural network classifier, and rule-based extraction, which we describe in the next section.

## Methods

The Institutional Review Board (IRB) of the University of Pennsylvania reviewed this study and deemed it to be exempt human subjects research under Category (4) of Paragraph (b) of the U.S. Code of Federal Regulations Title 45 Section 46.101 for publicly available data sources (45 CFR §46.101(b)(4)).

### Data collection

In previous work [10], we manually annotated more than 100,000 tweets in approximately 200 users' timelines, including reports of age. In the present study, we leveraged these annotations to develop handwritten, high-recall regular expressions—search patterns designed to automatically match text strings—to retrieve tweets that potentially mention a user's age between 10 and 99.

We deployed 26 regular expressions on two collections of public tweets: (1) more than 1.1 billion tweets (413,435,160 users) from the 1% Twitter Sample Application Programming Interface (API), collected between March 2015 and September 2019, and (2) more than 1.2 billion tweets posted by 245,927 users who have announced their pregnancy on Twitter [14]. One the one hand, the Twitter Sample API allows us to model the general detection of exact age based on the demographics of Twitter users. On the other hand, detecting the exact age of users who have announced their pregnancy on Twitter represents challenges that may be posed by more specific applications—for example, disambiguating the age of the user from the gestational age of the baby. After automatically ignoring retweets and removing "reported speech" (e.g., quotations, news headlines) [15], the regular expressions matched 1,340,015 tweets from the Twitter Sample API, and 997,486 tweets from the pregnancy collection.

## Annotation

To train and evaluate supervised machine learning algorithms, annotation guidelines were developed to help five annotators distinguish tweets that self-report a user's exact age ("age" tweets) from those that do not ("no age" tweets). For tweets that were annotated as "age," the annotators also identified the user's exact age that the tweet explicitly or implicitly reports. The annotators independently annotated a random sample of 11,000 of the 2,337,501 matching tweets—5,500 posted by unique users in each of the two collections. Among the 11,000 tweets, 10,000 were dual annotated, and 1000 were annotated by all five annotators. Based on the 1000 tweets that were annotated by all five annotators, the inter-annotator agreement for distinguishing "age" and "no age" tweets was 0.80 (Fleiss' kappa). For the "age" tweets on which the annotators agreed, the inter-annotator agreement for identifying the user's age was 0.95 (Fleiss' kappa). The first author of this paper resolved the class and age disagreements among the 11,000 tweets. Upon resolving the disagreements, 3543 (32%) of the tweets were annotated as "age," and 7457 (68%) as "no age." Table 1 presents examples of "age" (+) and "no age" (-) tweets.

**Table 1. Sample tweets manually annotated as "age" (+) or "no age" (-).**

|   | Tweet | Class | Age |
|---|---|---|---|
| 1 | It's my 21st birthday today. But who cares..... ITS FINALLY AUGUST!!!!! | + | 21 |
| 2 | It's crazy, tomorrow I'll be 20. I'm getting so OLD. | + | 19 |
| 3 | can't believe im going to be 21 .... i want to be a teenager again | + | 20 |
| 4 | I graduate in May only focusing on me and my child.. watch me at 21 | - | NA |
| 5 | Had just turned 18 then found out I was pregnant 2 weeks later | - | NA |

Tweets were annotated as "age" if the user's exact age could be determined, from the tweet, at the time the tweet was posted. In Table 1, Tweet 1 is a straightforward example of an "age" tweet, in which the user's exact age is explicitly stated. Although Tweet 2 does not explicitly state the user's age, it can be inferred from the fact that the user reports turning 20 tomorrow. Tweet 3 does not specify when the user will be 21, but it would be annotated as "age" under the assumption that the tweet is referring to the user's next birthday. Tweet 4, however, would be annotated as "no age" because it is ambiguous about whether the user was 21 when the tweet was posted, or whether the user is referring to a future age. Tweet 5 also would be annotated as "no age" because it is ambiguous whether the user was 18 when the tweet was posted, or whether the user is referring to age further in the past. Of course, tweets also would be annotated as "no age" if they obviously do not refer to the user or an age. Table 1 illustrates some of the challenges of training machine learning algorithms to automatically distinguish "age" and "no age" tweets.

## Classification

We used the 11,000 annotated tweets in experiments to train and evaluate supervised machine learning algorithms for binary classification of "age" and "no age" tweets. For the classifiers, we used the WLSVM Weka Integration of the LibSVM [16] implementation of Support Vector Machine (SVM), and two deep neural network classifiers based on bidirectional encoder representations from transformers (BERT): the BERT-Base-Uncased [17] and RoBERTa-Large [18] pretrained transformer models in the *Flair* Python library. We split the tweets into 80% (training) and 20% (test) random sets, stratified based on the distribution of "age" and "no age" tweets.

For the SVM classifier, we preprocessed the tweets by normalizing URLs, usernames, and digits, removing non-alphanumeric characters (e.g., punctuation) and extra spaces, and lowercasing and stemming [19] the text. Following preprocessing, we used Weka's default NGram Tokenizer to extract word n-grams (n = 1-3) as features in a bag-of-words representation. During training, each tweet was converted to a vector representing the numeric occurrence of n-grams among the n-grams in the training data. We used the radial basis function (RBF) and set the *cost* at $c = 32$ and the class weights at $w = 1$ for the "non-age" class and $w = 2$ for the "age" class, based on iterating over a range of values to optimize performance using 10-fold cross validation over the training set. We scaled the feature vectors before applying the SVM for classification.

For the BERT-based classifiers, we preprocessed the tweets by normalizing URLs and usernames, and lowercasing the text. After assigning vector representations to the tweet tokens based on the pretrained BERT model, the encoded representation is passed to a dropout layer (drop rate of 0.5), followed by a softmax layer that predicts the class for each tweet. For training, we used Adam optimization, 10 epochs, and a learning rate of 0.0001. During training, we fine-tuned all layers of the transformer model with our annotated tweets. To optimize performance, the model was evaluated after each epoch, on a 5% split of the training set.

## Extraction

We used the 2834 "age" tweets in the training set to develop a rule-based module that automatically extracts the exact age from "age" tweets. First, the module preprocesses the "age" tweets, including replacing spelled-out numbers with digits, removing URLs and usernames, which may contain digits, and removing spaces and other non-alphanumeric characters between digits (e.g., *the big 3-0*). Then, the module uses an optimized sequence of 87 handwritten regular expressions to match linguistic patterns containing two consecutive digits. Finally, the module applies a simple mathematical operation to the digits in the matching pattern, based on the regular expression that the tweet matches. If a tweet does not match one of the query patterns, the module simply extracts the first two-digit group from the tweet. Table 2 presents examples of matching patterns (bold) in "age" tweets and their associated age extraction rules.

**Table 2. Sample matching patterns (bold) in "age" tweets and their age extraction rules.**

|    | Tweet | Rule | Age |
|----|-------|------|-----|
| 1  | **Two more years until my 21st** birthday! Can't wait!  #surprise | 21 – 2 | 19 |
| 2  | **It's my 18th** birthday! And we have to go to school | 18 | 18 |
| 3  | excited for **my 18th** but also don't want to grow up | 18 – 1 | 17 |
| 4  | I started having #depression **20 yrs ago at the age of 19**. | 20 + 19 | 39 |
| 5  | I started **at 28 and I'm currently doing a PhD at 35**. | 28 < 35 | 35 |
| 6  | i feel like i'm going through a midlife crisis **at the age of 21** | 21 | 21 |
| 7  | I've **turned 21 three times** now. I don't think I can turn it a 4th. | 21 + (3 – 1) | 23 |
| 8  | I'm right there with you. Recently **turned 47**. | 47 | 47 |
| 9  | I was just reminded that I'm **turning 18 in 3 weeks** I feel old | 18 – (3 / 52) | 17 |
| 10 | I'm going out for the first time tonight **since turning 21** | 21 | 21 |

In Table 2, the users refer to *my* birthday in Tweet 1 (*my 21st*), Tweet 2 (*my 18th*), and Tweet 3 (*my 18th*), but the age extraction rule is different for each of these tweets based on the linguistic context in which this reference occurs. After preprocessing, the context in Tweet 1 includes an additional group of digits referring to a future time period (*two more years until*). Because the unit of time is *years*, the age is defined by simply subtracting the first group of digits (*2*) from the second (*21*). Fig 1 provides the Python code illustrating how our extraction module automatically identifies the age for Tweet 1. Tweet 4 includes an additional group of digits referring to a past time period (*20 yrs ago*), so the age is defined by adding the first group of digits (*20*) to the second (*19*). Tweet 5 also includes an additional group of digits referring to the past (*28*), but the reference is to a past age (*at 28*), rather than a time period, so the age is defined by the greater of the two groups of digits (*35*). Tweet 4, Tweet 5, and Tweet 6 illustrate how the age extraction rules vary depending on the specific pattern in which *at (the age of)* occurs.

**Fig 1. Sample Python code for extracting age from patterns with a unit of time.**

```python
regex_digits = re.compile(r"\d+")

if match := re.search(r"\b\d+\s*(more\s*)?(day|week|month|year|yr)s*\s*(to\s*go\s*)?
(until|till|til|away\s*from)\s*(my|our)\s*\d\d\s*(th|st|nd|rd)", tweet, flags=re.IGNORECASE):

    match_digits = regex_digits.findall(match.group())
    quantity = int(match_digits[0]) #e.g., 2 in "2 more years until my 21st"
    future_age = int(match_digits[1]) #e.g., 21 in "2 more years until my 21st"

    match_unit = re.search(r"(day|week|month|year|yr)", match.group(), flags=re.IGNORECASE)
    unit = match_unit.group()

    if unit.lower() == "day":
        units_in_year = 365
    elif unit.lower() == "week":
        units_in_year = 52
    elif unit.lower() == "month":
        units_in_year = 12
    elif unit.lower() == "year" or "yr":
        units_in_year = 1

    age = future_age - math.ceil(quantity / units_in_year)
```

Table 2 also illustrates the importance of optimizing the order in which the tweets match the query patterns. For example, Tweet 7 and Tweet 8 both report that the user *turned* an age, but if the pattern in Tweet 8 were applied before the pattern in Tweet 7, the age would be incorrectly extracted from Tweet 7 as 21. In addition, Tweet 7 illustrates that some patterns define the time period *(three times)* as the second group of digits (*3*), rather than the first (*21*)—a distinction that is especially important for age extraction rules that are based on subtraction, as in Tweet 9. In Tweet 9, however, the subtracted unit of time is *weeks*, rather than *years*, so, as Fig 1 illustrates, the time period would be converted to years by dividing the second group of digits (*3*) by 52, then rounding up to the nearest integer (1). While *turning* in Tweet 9 refers to a future age (*18*), it refers to a present age in Tweet 10 (*21*), in which case the age is extracted simply as the matching group of digits. Fig 2 illustrates our end-to-end pipeline.

**Fig 2. ReportAGE: an automatic natural language processing (NLP) pipeline for extracting the exact age of Twitter users based on self-reports in their tweets.**

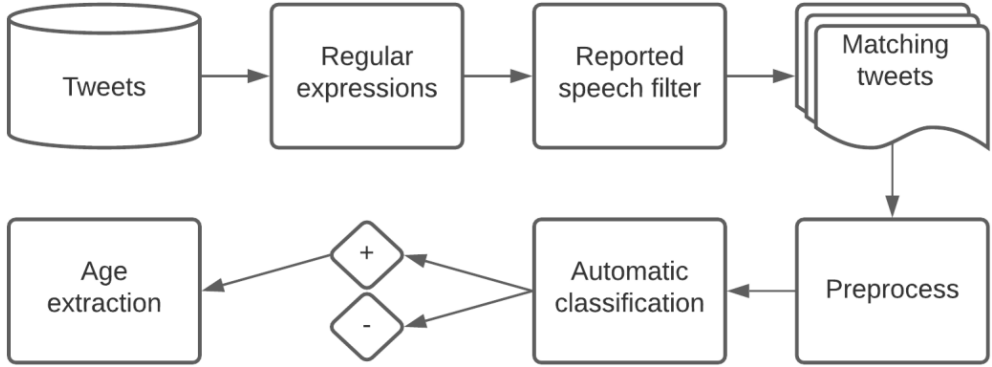

# Results and discussion

We evaluated an SVM classifier and two deep neural network classifiers on a held-out test of 2200 annotated tweets. For the "age" class, the SVM classifier achieved an $F_1$-score of 0.772 (precision = 0.734, recall = 0.814); the classifier based on the BERT-Base-Uncased pretrained model achieved an $F_1$-score of 0.879 (precision = 0.826, recall = 0.941); and the classifier based on the RoBERTa-Large pretrained model achieved an $F_1$-score of 0.914 (precision = 0.905, recall = 0.942), where:

$$F_1\text{-score} = \frac{2 \times recall \times precision}{recall + precision}; \quad recall = \frac{true\ positives}{true\ positives + false\ negatives}; \quad precision = \frac{true\ positives}{true\ positives + false\ positives}$$

For evaluating automatic classification, *true positives* are "age" tweets that were correctly classified; *false positives* are "no age" tweets that were incorrectly classified as "age;" and *false negatives* are "age" tweets that were incorrectly classified as "no age."

When the rule-based age extraction module was evaluated using the class predictions of the best-performing classifier (RoBERTa-Large), it achieved an $F_1$-score of 0.855 (precision = 0.805, recall = 0.914) for the "age" class. When it was evaluated directly on the annotated test set, it achieved an $F_1$-score of 0.931 (precision = 0.873, recall = 0.998) for the "age" class. For evaluating age extraction, *true positives* are "age" tweets that were automatically classified correctly and from which the age was extracted correctly (581 tweets); *false positives* are either "age" tweets that were automatically classified correctly but from which the age was extracted incorrectly (73 tweets), or "non-age" tweets that were automatically classified incorrectly as "age" and from which an age was extracted (68 tweets); and *false negatives* are either "age" tweets that were automatically classified correctly but from which no age was extracted (1 tweet), or "age" tweets that were automatically classified incorrectly as "non-age" (54 tweets). Table 3 presents examples of false positives and false negatives.

**Table 3. Sample false positive and false negative tweets for the "age" class, with their actual class (AC), predicted class (PA), actual age (AA), and predicted age (PA).**

|   | Tweet | AC | PC | AA | PA |
|---|---|---|---|---|---|
| 1 | Blessed to see my 22nd birthday! I feel good to be alive. | + | + | 22 | 21 |
| 2 | The most exciting part of turning 25 is that my insurance is dropping 20 bucks per month. | + | + | 25 | 24 |
| 3 | Got to love Facebook for reminding me of my 21st bday cruise | - | + | NA | 20 |
| 4 | Who will be going to two 21st birthdays next week and doesn't have anything nothing to wear?! ME | - | + | NA | 20 |
| 5 | Big 30 coming up on the 31st | + | - | 29 | NA |

For 61 (84%) of the 73 false positive tweets that were automatically classified correctly but from which the age was extracted incorrectly, the extracted age was only one year different from the annotated age—in particular, one year less for 43 (70%) of these 61 false positives. Among these 43 false positives, 31 (72%) of them matched age extraction patterns referring either to *my* birthday (19 false positives), as in Tweet 1 in Table 3, or *turning* an age (12 false positives), as in Tweet 2. Among the 68 false positives that were automatically classified incorrectly as "age," the digits in 58 (85%) of them do refer to an age. Among these 58 false positives, 21 (36%) of them self-report an age that the annotators had determined was temporally ambiguous, as in Tweet 3, and 14 (24%) of them that had been determined not to be self-reports do include a personal reference to the user elsewhere, as in Tweet 4. In contrast, 20 (37%) of the 54 false negatives that were automatically classified incorrectly as "age" do not explicitly refer to the user, as in Tweet 5.

We deployed ReportAGE end to end on the 1.2 billion tweets in our pregnancy collection [14]. In contrast to the Twitter Sample API, our pregnancy collection contains users' timelines (i.e., all of their tweets posted over time), which enables us to estimate the proportion of users for which ReportAGE can predict an exact age based on their tweets. Ignoring retweets, the regular expressions matched 1,128,754 tweets posted by 190,372 (77%) of the 245,927 users in the collection. After automatically removing 131,268 "reported speech" tweets [15], the regular expressions matched 997,486 tweets posted by 186,211 (76%) of the 245,927 users. We deployed the RoBERTa-based classifier on these 997,486 tweets, and 411,083 (41%) of them were automatically classified as "age" tweets. Finally, we deployed the age extraction on these 411,083 tweets, which predicted ages for 410,938 of them, posted by 132,637 (54%) of the 245,927 users.

The performance ($F_1$-score = 0.855) and coverage (54% of users) of ReportAGE suggest that our tweet-based approach can detect exact age for many more users than an approach based on extracting self-reports of age from their profiles [13]. Because ReportAGE deploys a classifier on only tweets that match regular expressions, it can scale to this large number of users without modeling hundreds or thousands of each user's posts [3-6, 8]. The regular expressions also help address the selection bias noted by Morgan-Lopez et al. [8] as a limitation of their study—that is, evaluating a model for detecting any user's age based only a population of users who announce their birthday in tweets. Because the regular expressions in ReportAGE are the same ones we used to collect the tweets in our annotated training and test sets, ReportAGE extracts ages only for users who have posted tweets that match patterns represented in our evaluation of performance.

Our evaluation of coverage, however, may reflect a selection bias towards users who are in an age group associated with pregnancy—that is, if users who are younger report their age on Twitter more often than that of users who are older. Deploying ReportAGE on our pregnancy collection also exemplifies a limitation of extracting age from tweets in users' timelines. Whereas deploying ReportAGE in real-time—for example, directly to the tweets collected from the Twitter Streaming API—would extract a user's present age, deploying ReportAGE on the tweets in a user's timeline may extract a past age. Normalizing a past age to the user's exact age in the present or another point in time, which is beyond the scope of this study, would be limited for users whose "age" tweets do not specify their birthday; however, if the past age were correctly extracted, the margin of error would be only one year, and this same limitation would apply to normalizing ages in profiles to points in the past. Directions for future work include developing methods for automatically aggregating the ages extracted from multiple tweets in a user's timeline.

## Conclusions

In this paper, we presented ReportAGE, an end-to-end NLP pipeline that automatically extracts the exact age of Twitter users based on self-reports in their tweets. The performance ($F_1$-score = 0.855) and coverage (54% of users) of ReportAGE demonstrate that individual tweets can be used to scale the automatic detection of exact age to a large number of users. Thus, ReportAGE can advance the utility of social media data for research applications that do not align with the predefined age groupings of binary or multi-class classification approaches, while overcoming the sparse reports of exact age in users' profiles. The performance of our age extraction module evaluated directly on the annotated test set ($F_1$-score = 0.931) suggests that the performance of ReportAGE can be improved with improvements to the performance of automatic classification.

## Acknowledgements

The authors would like to thank Karen O'Connor, Alexis Upshur, Isaac Valderrabano, and Saahithi Mallapragada for contributing to annotating the tweets; Jonathan Avila for contributing to the age extraction code and annotating the tweets; and Ivan Flores for contributing to database queries and the code for the end-to-end pipeline.

9. Culotta A, Ravi NK, Cutler J. Predicting the demographics of Twitter users from website traffic data. In: Proceedings of the Twenty-Ninth AAAI Conference on Artificial Intelligence; 2015. p. 72-78.

10. Golder S, Chiuve S, Weissenbacher D, Klein A, O'Connor K, Bland M, et al. Pharmacoepidemiologic evaluation of birth defects from social media postings during pregnancy. Drug Saf. 2019;42(3):389-400.

11. Davies SH, Langer M, Klein AZ, Gonzalez-Hernandez G, Dowshen N. Adolescent perceptions of menstruation on Twitter: opportunities for advocacy and education. J Adolesc Health. 2021;68(2):S9.

12. Flekova L, Carpenter J, Giorgi S, Ungar L, Preoţiuc-Pietro D. Analyzing biases in human perception of user age and gender from text. In: Proceedings of the 54th Annual Meeting of the Association for Computational Linguistics; 2016. p. 843-854.

13. Sloan L, Morgan J, Burnap P, Williams M. Who tweets? deriving the demographic characteristics of age, occupation and social class from Twitter user meta-data. PLOS One. 2015;10(3):e0115545.

14. Sarker A, Chandrashekar P, Magge A, Cai H, Klein A, Gonzalez G. Discovering cohorts of pregnant women from social media for safety surveillance and analysis. J Med Internet Res. 2017;19(10):e361.

15. Klein AZ, Cai H, Weissenbacher D, Levine LD, Gonzalez-Hernandez G. A natural language processing pipeline to advance the use of Twitter data for digital epidemiology of adverse pregnancy outcomes. J Biomed Inform X. 2020;8:100076.

16. Chang CC, Lin CJ. LIBSVM: a library for support vector machines. ACM Trans Intell Syst Technol. 2011;2(3):27.